\documentclass[10pt,journal,compsoc]{IEEEtran}

\usepackage{cite}
\usepackage{amsmath,amssymb,amsfonts}
\usepackage{graphicx}
\usepackage{textcomp}
\usepackage{xcolor}
\usepackage[numbers]{natbib}
\usepackage{algorithm}
\usepackage{algpseudocode}
\usepackage[colorlinks=true, urlcolor=blue, linkcolor=black, citecolor=black]{hyperref}
\ifCLASSINFOpdf
\else
\fi
\begin{document}
%
\title{Circular Clustering with Polar Coordinate Reconstruction}
%
%
%
%

\author{Xiaoxiao~Sun,~\IEEEmembership{Member,~IEEE,}
        and~Paul~Sajda,~\IEEEmembership{Fellow,~IEEE}
\IEEEcompsocitemizethanks{\IEEEcompsocthanksitem P. Sajda is with Department of Biomedical Engineering, Electrical Engineering, Radiology, and Data Science Institute, Columbia University, New York, NY, 10027.\protect \hfil\break
E-mail: ps629@columbia.edu
\IEEEcompsocthanksitem X. Sun is with Department of Biomedical Engineering, Columbia University, New York, NY, 10027.\protect \hfil\break
E-mail: xs2362@columbia.edu }
\thanks{ }}

\IEEEpubid{\begin{minipage}{\textwidth}\ \\[12pt] \centering Manuscript is under review in IEEE Transactions on Computational Biology and Bioinformatics\\
  1551-3203 \copyright 2023 IEEE. Personal use is permitted, but republication/redistribution requires IEEE permission.\\
  See http://www.ieee.org/publications standards/publications/rights/index.html for more information.
\end{minipage}} 

\IEEEtitleabstractindextext{%
\begin{abstract}
There is a growing interest in characterizing circular data found in biological systems. Such data are wide ranging and varied, from signal phase in neural recordings to nucleotide sequences in round genomes. Traditional clustering algorithms are often inadequate due to their limited ability to distinguish differences in the periodic component $\theta$. Current clustering schemes that work in a polar coordinate system have limitations, such as being only angle-focused or lacking generality. To overcome these limitations, we propose a new analysis framework that utilizes projections onto a cylindrical coordinate system to better represent objects in a polar coordinate system.  Using the mathematical properties of circular data, we show our approach always finds the correct clustering result within the reconstructed dataset, given sufficient periodic repetitions of the data. Our approach is generally applicable and adaptable and can be incorporated into most state-of-the-art clustering algorithms. We demonstrate on synthetic and real data that our method generates more appropriate and consistent clustering results compared to standard methods.  In summary, our proposed analysis framework overcomes the limitations of existing polar coordinate-based clustering methods and provides a more accurate and efficient way to cluster circular data.
\end{abstract}

\begin{IEEEkeywords}
Circular clustering, Polar Coordinate, Coordinate Reconstruction, Phase Synchronization, Circular DNA molecules 
\end{IEEEkeywords}}

\maketitle

\IEEEdisplaynontitleabstractindextext

%
\IEEEpeerreviewmaketitle

\ifCLASSOPTIONcompsoc
\IEEEraisesectionheading{\section{Introduction}\label{sec:introduction}}
\else
\section{Introduction}
\label{sec:introduction}
\fi

\IEEEPARstart{C}{lustering} is a common unsupervised learning technique for statistical analysis, used to explore interesting patterns inside a large datasets~\cite{anand2015effect}. It has been generally applied in many fields, including machine learning, data mining, pattern recognition, image analysis, and bioinformatics~\cite{merrell2016clustering}. During the past decade, a number of clustering algorithms have been proposed and categorized  based on their different characteristics (e.g., partitioning, hierarchical, density-based, etc.)~\cite{patil2019clustering,merrell2016clustering}. Corresponding distance measures are used to determine the similarity/dissimilarity between objects, with differences in these measures potentially resulting in  significantly different and inconsistent clustering outcomes~\cite{jain1999data}.

In terms of application, there is a growing interest in analyzing circular data from  different types of biological systems. For instance, the rhythmically synchronized activity in the brain is widely believed to play a fundamental role in coordinating information transfer during goal-directed behavior~\cite{fell2011role,canavier2015phase}. Recent studies suggest that the phase of ongoing oscillatory activity relative to endogenous or exogenous cues which will facilitate coordinated information transfer within circuits and between distributed brain areas and provide windows for optimal communication between two or more brain areas~\cite{frohlich2010endogenous,qasim2021phase,pantazatos2022functional}. Investigating phase synchronization (represented with polar coordinates) requires appropriate analysis tools to characterize aspects of the oscillatory activity, and one approach to do this is via unsupervised clustering. Another example is the analysis of round genomes that widely exist in living system, including circular RNA and extrachromosomal circular DNA molecules. These have been linked to multiple diseases including cancer~\cite{kristensen2019biogenesis,kim2020extrachromosomal}, and thus the clustering of CpG islands, gene locations and origin of replication can help discover active regions or hot spots~\cite{dong2019novel,govek2019clustering,debnath2021fast}. Additional  applications of circular clustering can also be found in iridology for disease diagnosis~\cite{hussein2013assessment}.      

In considering how clustering methods can be applied to oscillatory/circular data, it is important to note that most state-of-the-art clustering algorithms are not constructed in a way that is based on a polar coordinate representation. Although there are some circular clustering tools, they either only emphasize a polar angle difference (ignore the amplitude) or are designed with loss of generality -- i.e. they are designed to work on datasets that obey specific assumptions~\cite{merrell2016clustering,patil2019clustering,debnath2021fast}. 

In this paper, we propose a novel method which results in a  polar coordinate representation that is directly applicable to most cluster methods that are not specifically designed for analysis of circular data. Specifically, our approach maps polar coordinates to the lateral surface of cylindrical coordinates, creating a 3-dimensional representation that records differences in both distance $r$ from the origin and the angle $\theta$. This representation is then  unraveled and flattened into a rectangular plane which simplifies and generalizes the geodesic, in 3-dimensional space, to a Euclidean distance between two points in a 2-dimensional space. This reconstructed coordinate can then be used as an input to most clustering algorithms. We address the issue of circularity/periodicity in data by searching for clusters across multiple identical periods with repetitions of the original period. We present, as examples, methods that apply to partitioning (e.g., $K$-means), density-based (e.g., Density-Based Spatial Clustering of Applications with Noise (DBSCAN)), and hierarchical clustering (e.g., dendrogram) methods. Lastly, we demonstrate the utility of using this clustering approach for synthetic data and real biological data (e.g., phase synchronization of neural data and circular DNA molecules).  
     
\section{Related Work}
The main objective of circular clustering algorithm is to divide a set of data points into groups or clusters that are arranged in a circular or ring-shaped pattern~\cite{merrell2016clustering,patil2019clustering}. The circular pattern can be particularly useful in situations where the data is arranged in a circular or periodic manner, such as in the analysis of time series data, biological data, or geographic data~\cite{jain1999data,mardia2000directional}. 

One of the earliest and most well-known circular clustering algorithms is the $K$-means algorithm. This algorithm partitions the data into a pre-specified number of clusters, where each cluster is represented by its centroid~\cite{govek2019clustering}. While $K$-means is a powerful clustering algorithm, it has some limitations when it comes to circular clustering. For instance, $K$-means usually uses Cartesian distance as its distance metric, which may not be appropriate for circular data~\cite{mardia2000directional,charalampidis2005modified}.

To overcome these limitations, several variations of the $K$-means algorithm have been proposed~\cite{mardia2000directional,charalampidis2005modified,debnath2021fast}. For instance, the $K$-means clustering algorithm for circular data (KCC) uses a polar coordinate system to represent the data, where the distance between two data points is measured along the circumference of a circle~\cite{mardia2000directional}. The algorithm then adjusts the centroids based on the average angle and radius of the data points in each cluster. Circular $K$-means (CK-means) is another $K$-means based algorithm where it creates cluster vectors to contain directional information in a circular-shift invariant manner~\cite{charalampidis2005modified}. 

Despite their usefulness, $K$-means based algorithms have certain limitations, such as their sensitivity to initialization, the need for prior knowledge of the number of clusters, and difficulties with unequal cluster sizes. As a result, alternative circular clustering algorithms that do not rely on $K$-means have been developed, including mean shift-based~\cite{chang2012mean} and learning-based clustering methods~\cite{abraham2013unsupervised,li2018adaptive,wang2021adaptive}. However, it is important to note that the performance of mean shift-based methods can be influenced by the bandwidth parameter of the circular kernel function, which controls the smoothing of the density function. Moreover, learning-based clustering methods require labeled data for training, which can be expensive and time-consuming to obtain. In addition, these methods perform suboptimally when confronted with complex data distributions that are not consistent with the model assumptions. Lastly, as the models employed can be quite complex with numerous parameters, interpretation of the results can be difficult.

Overall, circular clustering is a valuable approach for finding patterns in circular data, though the existing methods have significant limitations that make them unsuitable for many applications. Therefore, there is a need for novel and versatile algorithms that can model circular patterns explicitly across a broad range of applications. Such algorithms would help to overcome the limitations of current circular clustering methods and enable more accurate and efficient analysis of circular data.

\section{Methods}

\subsection{Distance Metrics on Cartesian Coordinates}
\vspace{-0.2em}
Similarity or dissimilarity measurements in clustering algorithms play an essential role in grouping or separating data~\cite{jain1999data}. A key step is to select suitable distance metrics. The simplest measure is the Manhattan distance (i.e., $L^1$ norm) which is equal to
the sum of absolute distances for each variable. The formula for this grid-like path distance $D$ between $X 
= (x_1,x_2\cdots,x_n)$ and $Y= (y_1,y_2\cdots,y_n) \in \mathbb{R}^n$ is:

\begin{equation}
    D(X,Y) = \sum_{i=1}^n|x_i - y_i|
\end{equation}

A more commonly used metric in many clustering algorithms is the Euclidean distance (i.e., $L^2$ norm or Euclidean norm) \cite{merrell2016clustering}. It is computed by taking the square root of the sum of the squares of the differences between each variable. The formula for this, as the-crow-flies, distance $D$ between $X$ and $Y$ is:

\begin{equation}
    D(X,Y) = \Big(\sum_{i=1}^n(|x_i - y_i|)^2\Big)^{\frac{1}{2}}
\end{equation}

\subsection{Clustering on Polar Coordinates}
\begin{figure*}[h!]
    \centering
    \includegraphics[width=1\linewidth]{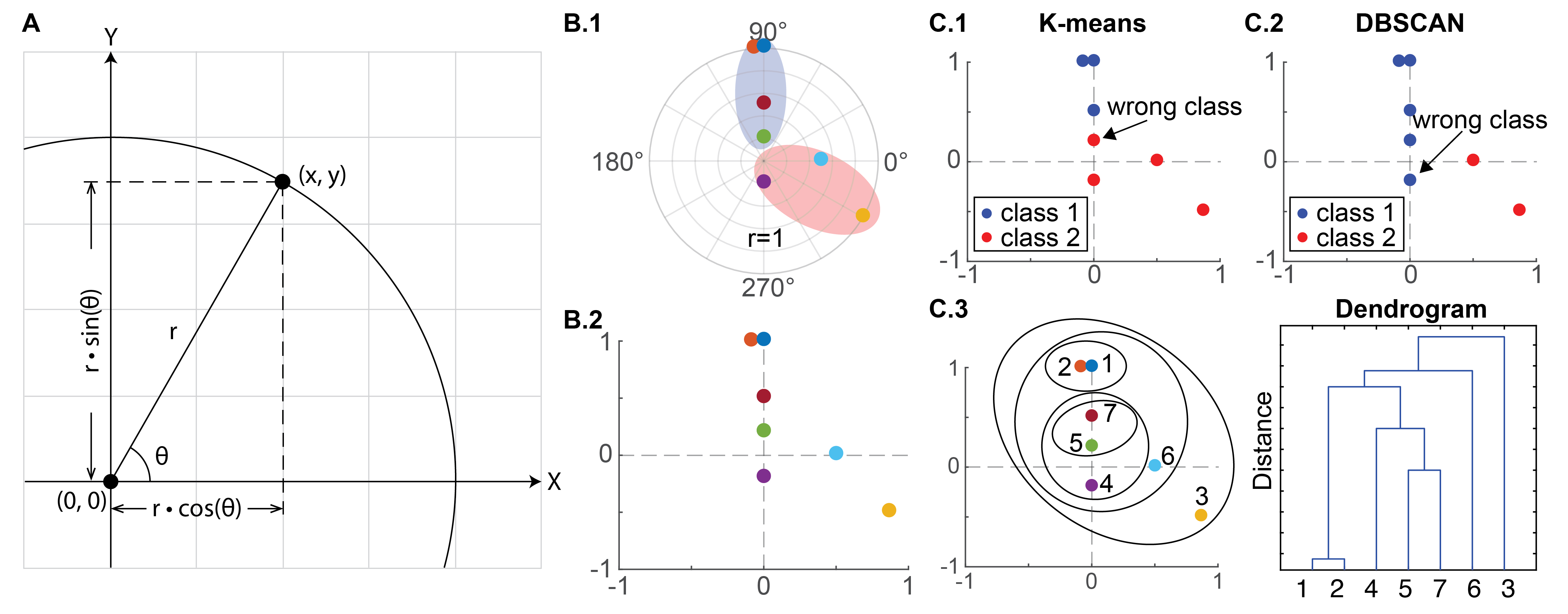}
     \caption{Common approach to perform clustering analysis for data specified in polar coordinates. (A) Illustration converting from polar coordinates $(r, \theta)$ to Cartesian coordinates (x,y) as~\eqref{eq:polartocart}. (B.1) Seven example sample points in polar coordinates are shown for illustration purposes (points in the same shaded area belong to the same class). (B.2) Transforming points in (B.1) to Cartesian coordinates (points correspondences indicated by color). (C.1) Clustering result generated by $K$-means method ($K=2$) where the green point $(r=0.2,\theta=\pi/2)$ in (B.1) is incorrectly classified. (C.2) Clustering result generated by DBSCAN method (neighborhood search radius $\epsilon=0.5$ and minimum number of neighbors $n_{\min}=2$) where the purple point $(r=0.2,\theta=3\pi/2)$ in (B.1) is incorrectly classified. (C.3) Clustering result generated by the hierarchical clustering method. The figure on the left graphically illustrates the way the linkages group the objects into a hierarchy of clusters, and the figure on the right is the corresponding dendrogram of clusters where the x-axis corresponds to the leaf nodes of the tree, and the y-axis corresponds to the linkage distances between clusters. Similar to the $K$-means and DBSCAN methods, incorrect grouping arise for the sample points that are labeled as 4, 5 and 7.}
    \label{fig:example}
\end{figure*}

A general approach for solving clustering problems in polar coordinates is to convert the representation of the variables from polar coordinates $(r,\theta)$ to Cartesian coordinates $(x,y)$ (see \eqref{eq:polartocart} and Fig.~\ref{fig:example}A) and calculate the Euclidean distance with Cartesian coordinates,

\vspace{-0.5em}
\begin{equation}
    (x,y) = (r\cos(\theta)
    ,r\sin(\theta)) \label{eq:polartocart}.
\end{equation} 
However, since polar coordinates have a circular element $\theta$, clustering methods using the Euclidean distance can lead to an incorrect result. For instance, if two points have a $180^{\circ}$ phase difference but are close to the origin, any $L^2$-based clustering method would give a wrong conclusion because of its limited ability to distinguish differences in $\theta$ (i.e., these two misclassified points with the same radius but $180^\circ$ out-of-phase in Fig.~\ref{fig:example}(B.1) can also be interpreted as in-phase but having  opposite amplitudes in Fig.~\ref{fig:example}(B.2)). Furthermore, a clustering method with Cartesian coordinates may not be able to extract corresponding meaningful information in polar coordinates. While several circular clustering algorithms have previously been proposed to address these issues, most of them only differentiate the $\theta$ component \cite{merrell2016clustering,debnath2021fast} or are extremely complex \cite{li2018adaptive,patil2019clustering}, leaving researchers with no direct or generally applicable clustering tools optimized for data in polar coordinates.

\subsection{Polar Coordinate Reconstruction}
Since Euclidean distance is not a suitable metric for data in polar coordinates, one straightforward solution is to use other distance metrics (e.g., ArcCosineDistance, Canberra
Distance, Lorentzian
Distance, etc.)~\cite{jain1999data,merrell2016clustering}. However, this requires researchers to recreate different clustering algorithms specialized for their data, though this results in a  loss of generality. Moreover, such methods may still fail to capture the differences in both $r$ and $\theta$  (e.g., most circular clustering algorithms emphasize $\theta$~\cite{debnath2021fast}). Alternatively, one can convert a polar representation of points to a non-cartesian representation, though it is challenging to deal with the periodic element ($\theta = \theta + 2k\pi,k\in \mathbb{N}$) and its circular characteristic ($ \Delta \theta = \theta_1 - \theta_2$ or $ \Delta \theta = \theta_2 - \theta_1 $, i.e., the angle difference between two points can be approached either clockwise or counterclockwise). Additionally, these converted coordinates may not be well-suited for   state-of-the-art clustering methods. Thus, there is a need for a circular clustering framework which is generalizable across clustering approaches and enables one to consider both magnitude and phase angle in the distance metric.

\subsubsection{Proposed Coordinate System}
\begin{figure}[h!]
    \centering
    \includegraphics[width = 1\linewidth]{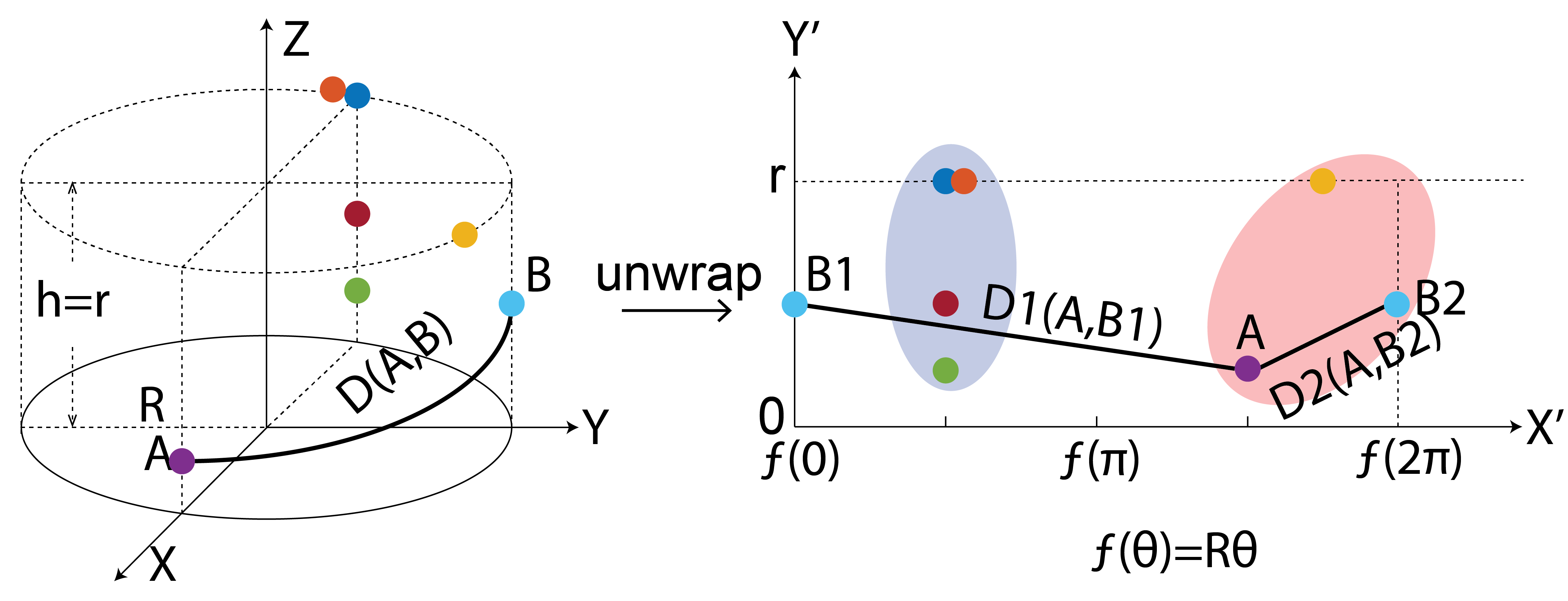}
    \caption{Reconstruction of polar coordinates on the rectangular plane $(X',Y')$ using cylindrical coordinates. Mapping the points in polar coordinates to the lateral surface of the cylinder where the height is equal to the radius of the polar coordinates system ($h=r$) and the base circle has radius $R$. Values of $R$ control the weights of $\theta$ as described in \eqref{eq:reccoordinate}. By flattening the lateral surface, we  obtain a 2-dimensional rectangular space where the $X'$-axis represents $\theta$ and $Y'$-axis represents $r$. The length of the $X'$-axis equals the circumference of the base circle (i.e., $2\pi R$). }
    \label{fig:cylinder}
\end{figure}

Since cylindrical coordinates are an extension of 2-dimensional polar coordinates to 3-dimensions, we propose to use such a coordinate system to capture dissimilarities in both $r$ and $\theta$ . After mapping points in polar coordinates to the lateral surface of the cylinder with \eqref{eq:cylinder}, the distance between two points on the lateral surface  (i.e., geodesic) can be estimated as the Euclidean distance  on the rectangular space $(X',Y')$. This rectangular space is obtained by unraveling the cylinder and flattening it into a plane (see Fig.~\ref{fig:cylinder}, where the lateral surface area of a cylinder is the area of a rectangle). Any point $(x',y')$ on $(X',Y')$ can be written as \eqref{eq:reccoordinate},
\begin{align}
    (x,y,z) = (R\cos(\theta),R\sin(\theta),r) \label{eq:cylinder} \\
    (x',y') = (f(\theta),r)=(R\theta,r) \label{eq:reccoordinate}
\end{align}
where $R$ is the radius of the base circle of the cylinder. $R$ is a parameter that can be adjusted based on the weights (importance) of difference in $\theta \in [0,2\pi)$. Greater $R$ indicates the difference observed in $\theta$ is more important than $r$.   

\subsubsection{Solving for Circularity}
\label{sec:searching}
As shown in the rectangular space of Fig.~\ref{fig:cylinder}, because point $B$ on the cylindrical coordinates can be expressed as either $ B1=(0,r)$ or $B2=(2\pi,r)$, then the distance between example point $A$ and $B$ is given by  either $D1(A,B1)$ or $D2(A,B2)$. In this example, the appropriate distance to generate the correct clustering result is $D2(A,B2)$, so it is essential to locate the accurate distance that moves clockwise or counterclockwise in a circle. Inspired by the coordinate optimization approach in phase-controlled robotics \cite{lin2020optimizing}, we can take advantage of the periodic nature of the polar coordinate system. Specifically, we address the circularity problem by leveraging its periodic nature through period repetition and explore three different clustering methods to demonstrate how this technique can be used.

\begin{figure}
    \centering
    \includegraphics[width=1\linewidth]{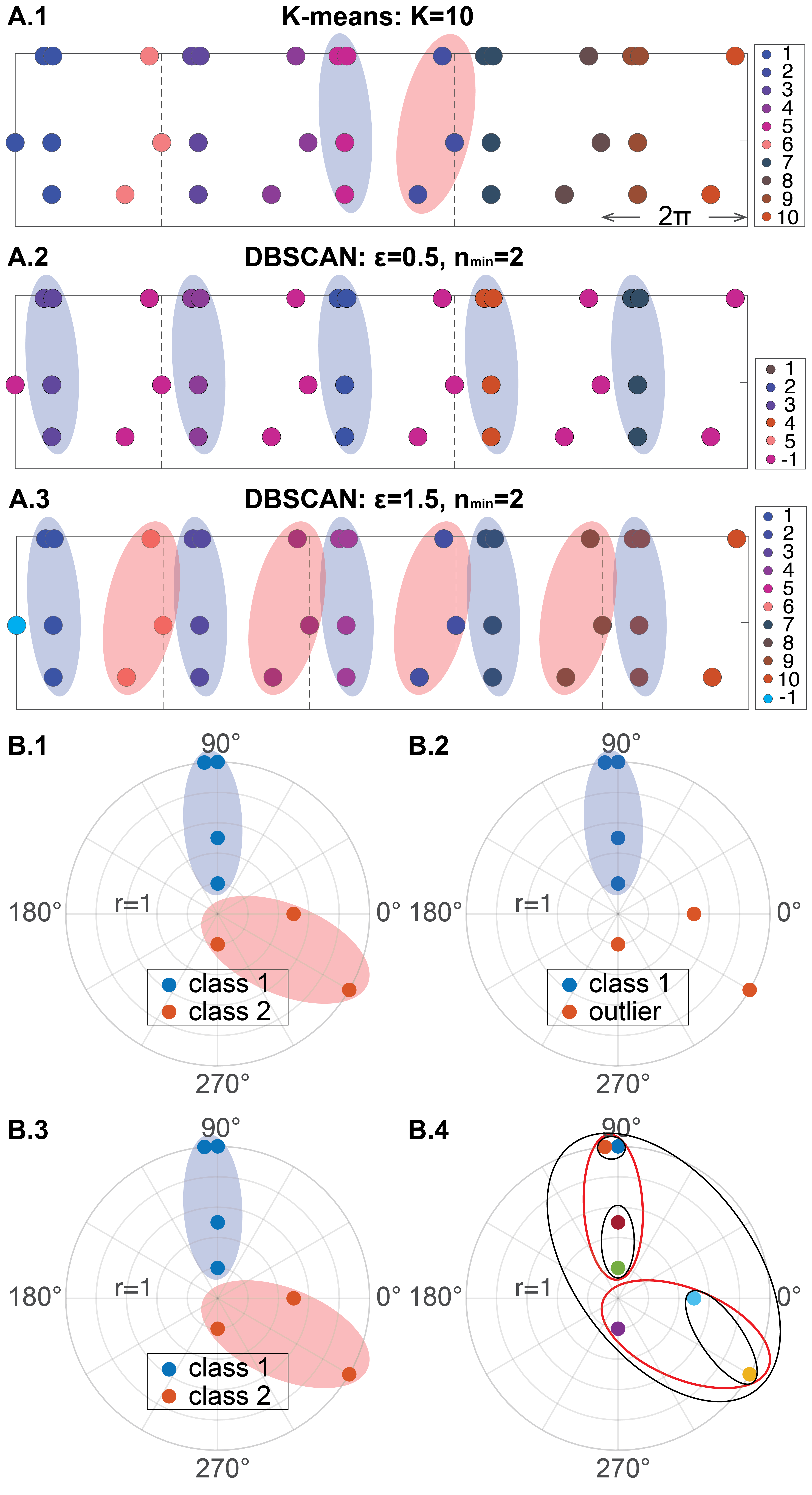}
    \caption{Solving circularity with period repetition for $K$-means, DBSCAN and hierarchical clustering methods. (A.1) In the given example, with four period repetition, the question of finding two classes within one period becomes the question of finding $10=2\times (4+1)$ classes within the total $5=4+1$ periods with the $K$-means method. Points with the same color belong to the same class. We can locate the correct cluster output in the middle (i.e., 3rd) period. (A.2) We apply the same DBSCAN parameters as Fig.~\ref{fig:example} ($\epsilon=0.5$ and $n_{\min}=2$) on the rectangular plane with five periods. Points with the same color belong to the same class. We consistently identify the same pattern for class 1 (four points are classified into the same class repetitively five times, highlighted as blue shadow). All other points that are unclassified are defined as outliers, which are also the outliers defined by given parameters (referred as label `-1'). (A.3) We apply DBSCAN with different parameters ($\epsilon=1.5$ and $n_{\min}=2$) on the rectangular plane with five periods. Two patterns have been identified this time and those two patterns include all 7 sample points, so two classes are found with no outliers. (B.1) Clustering result of $K$-means in (A.1). (B.2) Clustering result of DBSCAN in (A.2). (B.3) Clustering result of DBSCAN in (A.3). (B.4) Clustering result of hierarchical clustering method with proposed algorithm, its result is improved from result shown in Fig.~\ref{fig:example}~C.3, and it result is consistent with  the $K$-means and DBSCAN methods where both class 1 and class 2 are included as sub-class (highlighted in red).} 
    \label{fig:stim7}
\end{figure}

\textbf{$K$-means:} Because of the periodicity of points in polar coordinates, the layout of points within each extended period (from $0$ to $2\pi$) in the rectangular plane would be identical. Suppose the number of classes is predetermined as $K$. In that case, the question of finding $K$ classes within one period is equivalent to the question to find $(c+1)\times K$ classes in the total extended rectangular plane when we repeat the period an additional $c$ times (i.e., $c+1$ periods in total). Based on the similarity of the extended periods, one can always find the correct clustering results to the problem in the middle period if we repeat the periods enough times (see Fig.~\ref{fig:stim7}A.1 and Algorithm~\ref{alg:kmeans}). Note: one repeats an even number of periods to keep the periods, other than the middle periods, nearly identical, e.g., $c=2n,n=1,2,\cdots$).    

\begin{algorithm}
\caption{Search algorithm for $K$-means}\label{alg:kmeans}
\begin{algorithmic}[1]
\Require $c=2n, n\in \mathbb{Z}^+$, clustering class $K$, clustering result of $2n+1$ periods $K_t$
\Ensure Final clustering result identified $K_{c}$
\State locate middle period: $N \gets n+1$
\State sample points within $N$: $P_{n+1} \in 360^{\circ}\times[n,n+1]$
\State corresponding clustering class for $P_{n+1}$: $K_{n+1}\subset K_t$
\State combination of $K_{n+1}$ based on number $K$: $C_K$
\For{$x =$ 1 to size($C_K$)}
\If{combination $C_x$ includes all the original sample points $P_1$ without repetition}
    \State $K_c \gets C_x$ 
\EndIf
\EndFor
\If{No such $C_x$ exists}
    \State increase the repetition number $c$ 
\EndIf

\end{algorithmic}
\end{algorithm}

\textbf{DBSCAN:} Repeating periods also works when the number of classes is unknown (e.g., clustering with density-based methods). Adapting the same parameters ($\epsilon$ and $n_{\min}$) used for the original period, we repetitively classify points belonging to the same group into one group (see Fig.~\ref{fig:stim7}A.2). Those points within those identical classes must be classified into one group (see Algorithm~\ref{alg:dbscan}). For example, in Fig.~\ref{fig:stim7}A.2, one type of class has been repetitively found. Another instance with different parameters is shown in Fig.~\ref{fig:stim7}A.3 and Fig.~\ref{fig:stim7}B.3, where two classes with no outliers are identified. Fig.~\ref{fig:stim7}B.3 provides the same conclusion as $K$-means method when $K=2$ in Fig.~\ref{fig:stim7}B.1. Period repetition does not change the definition of outliers in the DBSCAN algorithm, instead they are completely determined by the choice of $\epsilon$ and $n_{\min}$. More repetitions will provide higher confidence in the clustering result since the `correct' pattern would always be detected after repetition. 

\begin{algorithm}
\caption{Search algorithm for DBSCAN}\label{alg:dbscan}
\begin{algorithmic}[1]
\Require $c, c\in \mathbb{Z}^+$, clustering result of $c+1$ periods DB$_t$
\Ensure Final clustering DB$_{c}$
\State number of class has been identified: $k \gets$ unique(DB$_t$) 
\State samples points within each class $P_{k}$
\State map all points in $P_{k}$ to the original period $[0^{\circ}, 360^{\circ})$
\State count the class has the same points $Z_k$
\State setup the repetition threshold $Y$ \Comment{based on $c$}  
\For{$x =$ 1 to $k$}
\If{$Z_x \geq Y$}
    \State class $x$ should be included in DB$_c$ 
\EndIf
\EndFor
\State points not included in DB$_c$ are outliers 
\end{algorithmic}
\end{algorithm}

\textbf{Hierarchical Clustering (dendrogram): }  Period repetition can also be used in the hierarchical clustering method to find the correct minimal pairwise distance. For example, in the given samples in Fig.~\ref{fig:cylinder}, the distance between $A$ and $B$ would be $D_2$ which will be calculated when we repeat $B_1$ as $B_2$ in the second period, and the final hierarchical clustering result with reconstructed coordinates is given in Fig.~\ref{fig:stim7}~B.4. The details of this algorithm can be found in Algorithm~\ref{alg:hier}.      

\begin{algorithm}
\caption{Search algorithm for Hierarchical Clustering}\label{alg:hier}
\begin{algorithmic}[1]
\Require $c, c\in \mathbb{Z}^+$, all points $P$ within $c+1$ period 
\Ensure Final dendrogram HC$_{c}$
\State calculate pairwise distance by reconstructed coordinates: $Y \gets$ pdist($P$) 
\State pairwise combination within all sample points $P$: $C \gets$ nchoosek(1:size($P$),2)   
\State pairwise combination within the original sample points $P_1$: $C_1 \gets$ nchoosek(1:size($P_1$),2)   
\State map pairwise combination $P$ to the original period: $P' \gets \mod$($P$,size($P_1$))
\State Find the minimal pairwise distance calculated 
\For{$x =$ 1 to size($C_1$)}
\State $Y_{C_1} \gets$ $\min$($Y_x$)  \Comment{pairwise distance of $C_1$}
\EndFor
\State build the dendrogram by $Y_{C_1}$: $Z\gets$ linkage($Y_{C_1}$)     
\end{algorithmic}
\end{algorithm}

\section{Results}

In the previous section, we have shown that the proposed method  provides the correct classification for the given example with the $K$-means, DBSCAN, and hierarchical clustering methods (see Fig.~\ref{fig:stim7}). Next, we provide additional evidence of the value of the approach on both synthetic and real data.  

\vspace{-0.5em}

\subsection{Application to Synthetic Data}
\vspace{-0.3em}
The previous example only included seven points to illustrate the problem and introduce the framework. Now, we examine the clustering performance under different distribution scenarios. 
\subsubsection{Larger sample size}\label{sec:large}
Fifty points were randomly generated on the unit circle ($r=1$) based on two classes. In Fig.~\ref{fig:synthetic}A, we show how different $R$ (introduced in~\eqref{eq:reccoordinate}) would shape the clustering output with the $K$-means method. Smaller $R$ gives more balanced weight between $r$ and $\theta$, while larger $R$ favors angle-driven (emphasize on differentiating $\theta$) clustering results. The flexibility in the approach enables one to incorporate prior knowledge about the relative importance of both dimensions.  Note that a balanced weight for $R$ generates the same two classes from the underlying generative model and the angle-driven approach produces result identical to conventional $K$-means angle-focused circular clustering algorithms (e.g., Fast Optimal Circular Clustering (FOCC), brute force optimal circular clustering (BOCC), and heuristic circular clustering (HEUC)~\cite{debnath2021fast}, see Fig.~\ref{fig:supp_synth}A.1 in Appendix~\ref{sec:heir}).    

\subsubsection{Multiple classes ($K \geq 3$)}\label{sec:multi}
\begin{figure}[h!]
    \centering
    \includegraphics[width=1\linewidth]{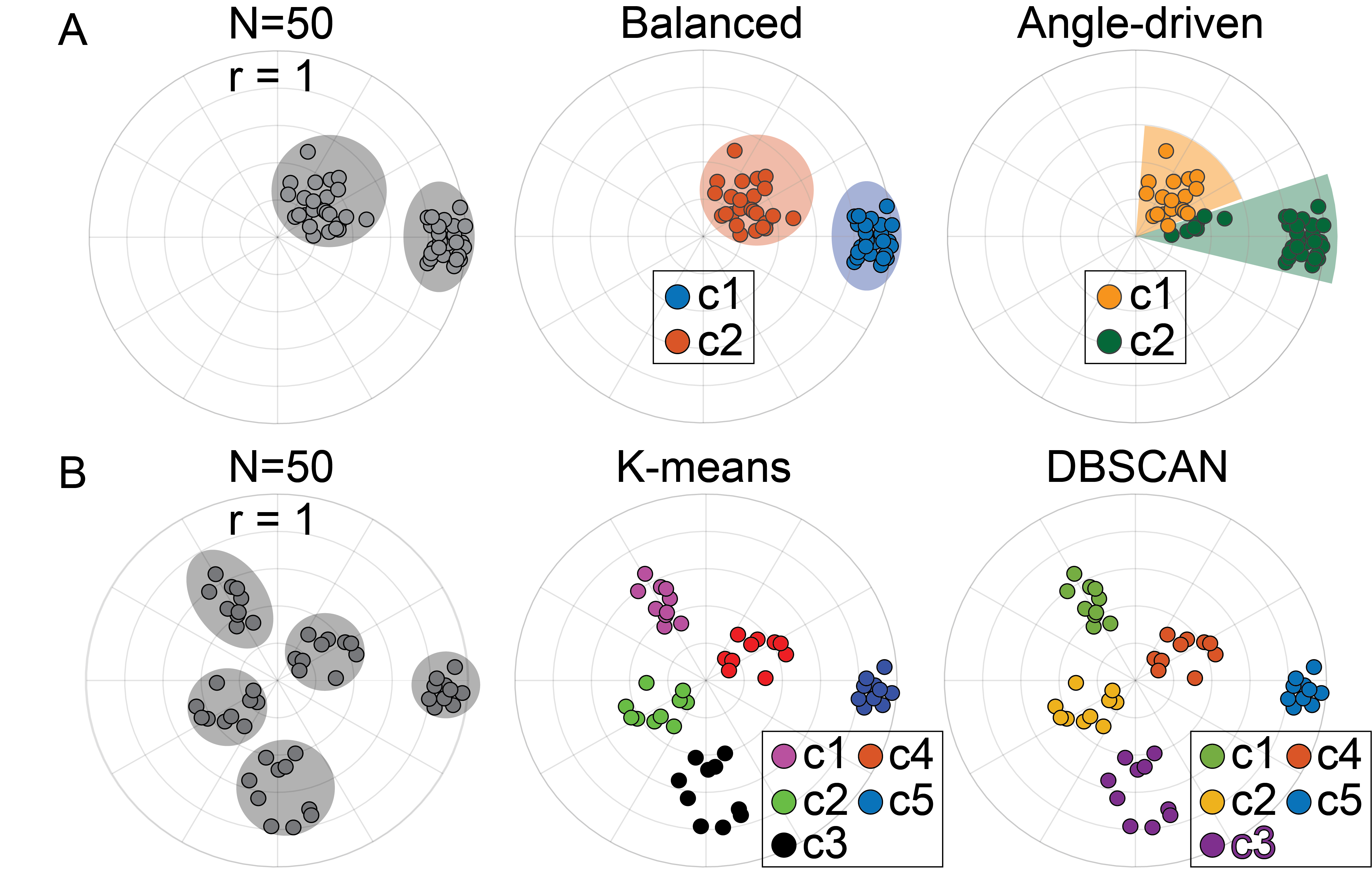}
    \caption{Application to synthetic data ($N=50$). (A). Clustering results with $K$-means on two groups. Firstly, 50 points are randomly generated based on two classes on the unit cycle (highlighted with the gray shadow). With a smaller $R$ applied, the clustering results have a balanced weight between $r$ and $\theta$, while angle-driven clustering results will be generated when a larger $R$ is used. The angle-driven clustering result is identical as the result generated by FOCC, BOCC and HEUC~\cite{debnath2021fast}. (B) Clustering results of multiple groups with $K$-means and DBSCAN. Another 50 points are randomly generated based on five classes (highlighted with the gray shadow) on the unit cycle. Applying both $K$-means and DBSCAN on our reconstructed representation yields the ground-truth as output. }
    \vspace{-0.5em}
    \label{fig:synthetic}
\end{figure}

During the previous discussion, only considered only clustering problem with two default classes. Here we use our  proposed method on a dataset generated based on five classes. More period repetitions are needed to cluster a larger number of classes.  Simulation results show that  both $K$-means and DBSCAN uncover the same clusters as the  generating  distribution (see Fig.~\ref{fig:synthetic}B). Additionally, for hierarchical clustering, the clades provide the same information as $K$-means and DBSCAN, where five clades are formed that  include ten points within each class  (see Fig.~\ref{fig:supp_synth}A.2 in Appendix~\ref{sec:heir}). 

\vspace{-0.8em}

\subsection{Application to Neural Data}


In a previous clinical study we measured the inter-trial phase coherence (ITPC) of the quasi-alpha oscillation (6-13Hz) from participants' EEG signals~\cite{ faller2022daily,pantazatos2022functional,sun2023daily}. In this study, we hypothesized that stronger and more precise inter-trial coherence would be a biomarker of the efficacy of a neurostimulation therapeutic. By mapping participant's post-stimulation quasi-alpha phase onto the unit circle and calculating the circular mean, we obtain 30 points (i.e., 30 experimental sessions) in polar coordinates. In this representation a  larger  $r\in [0,1]$ indicates consistency of each session's phase synchronization across all trials (greater $r$ means better synchronization) and $\theta$ refers to the phase of the synchronization (more details are available in \cite{faller2022daily}). We applied our proposed method to these data to assess the phase synchronization among all experimental sessions for each subject.

\begin{figure}[h!]
    \centering
    \includegraphics[width=1\linewidth]{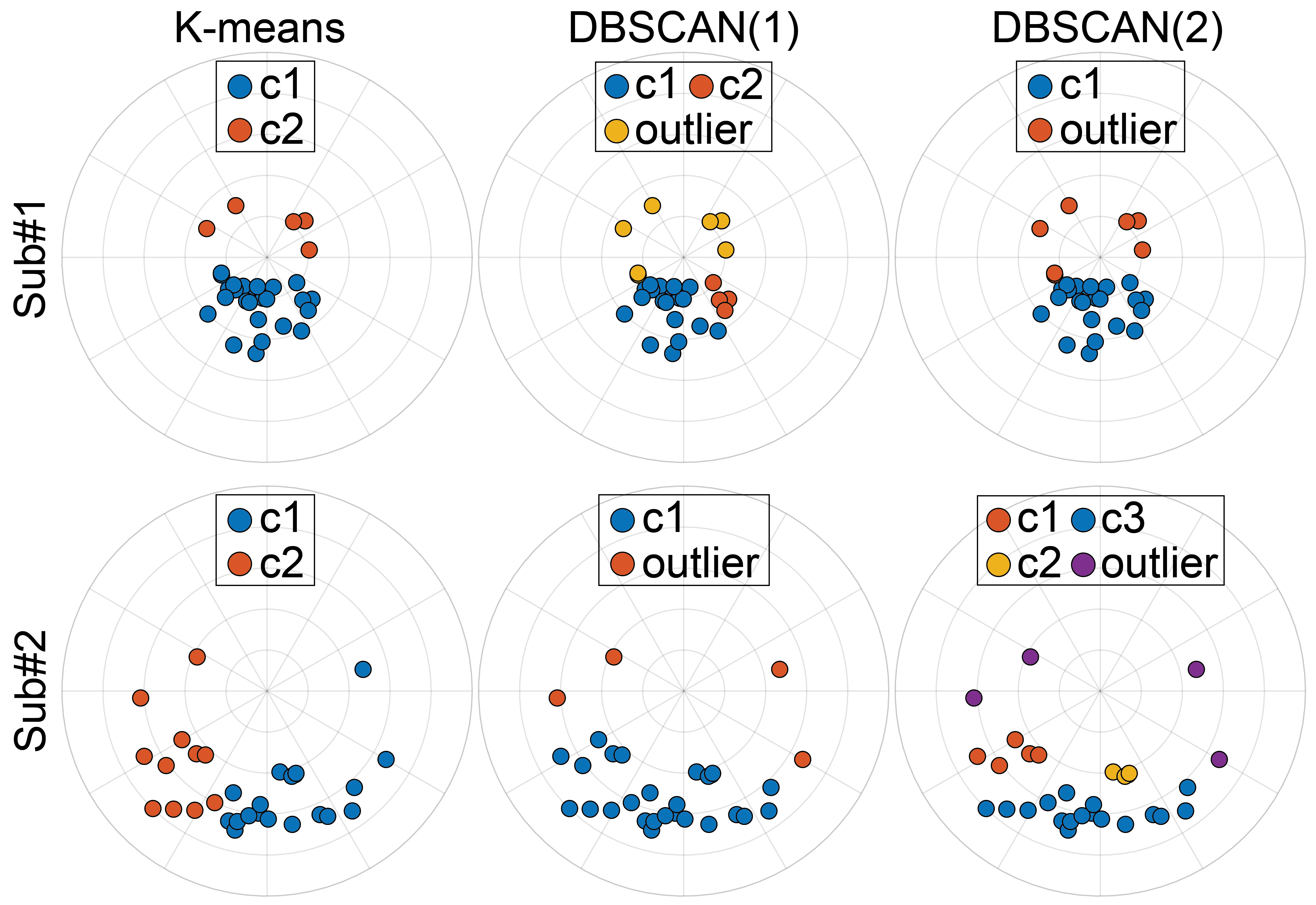}
    \caption{Application to neural data ($N=30$). Two subjects are used as examples. The output of $K$-means with a larger $R$ to capture the difference in $\theta$ is identical to the output of other angle-focused circular clustering method (FOCC with $K=2$, see Fig.~\ref{fig:supp_synth}B.1 in Appendix~\ref{sec:heir}). Additionally, DBSCAN with two sets of parameters is examined. DBSCAN(1) is generated with $\epsilon=0.25$ and $n_{\min}=5$, and DBSCAN(2) is generated with $\epsilon=0.2$ and $n_{\min}=3$ for both subjects. The DBSCAN clustering result of Sub\#1 shows this subject has consistent similarity within blue points (c1 group in (1) and (2)), while no such observation is seen in Sub\#2.}
    \label{fig:realdata}
\end{figure}

As shown in Fig.~\ref{fig:realdata}, when assuming the number of classes is predetermined ($K=2$, inlier and outlier), the angle-focused $K$-means method yields the same clustering result as the FOCC method~\cite{debnath2021fast} for both subjects. Moreover, the representation can also be used with DBSCAN when density based methods are desirable (see Fig.~\ref{fig:realdata}). Interestingly, the clustering result of Sub\#1 shows consistent similarity within blue points (similar class `c1' is defined with angle-focused $K$-means, DBSCAN(1), DBSCAN(2), and balanced $K$-means in Appendix~\ref{sec:heir}), which is consistent with our hypothesis and clinical experimental results--i.e. Sub\#1 is a subject who showed greater phase synchronization and better clinical improvement from the neurostimulation treatment.

\subsection{Application to DNA Data}
Another common application of circular clustering in biology is the analysis of DNA or RNA sequences. The circular representation of DNA or RNA sequences allows for the visualization of the periodicity or cyclic patterns that may be present. Clustering techniques can be applied to these circular sequences to identify similar patterns or motifs, which can provide insights into the structure and function of the sequences~\cite{qi2009classification,kristensen2019biogenesis}. With the polar coordinate representation of DNA sequence proposed by Dai et al.~\cite{dai2012sequence} (or other methods to construct circular structure~\cite{kristensen2019biogenesis,kim2020extrachromosomal}), we can map any given biological sequence into polar coordinates where the radius and angles are determined by the distribution of the dual nucleotides (e.g., `AC', `AG', etc). Then by applying hierarchical clustering, we can gain insight into the relationships between different sequences and patterns in the sequence distribution.

\begin{figure}[h!]
    \centering
    \includegraphics[width=1\linewidth]{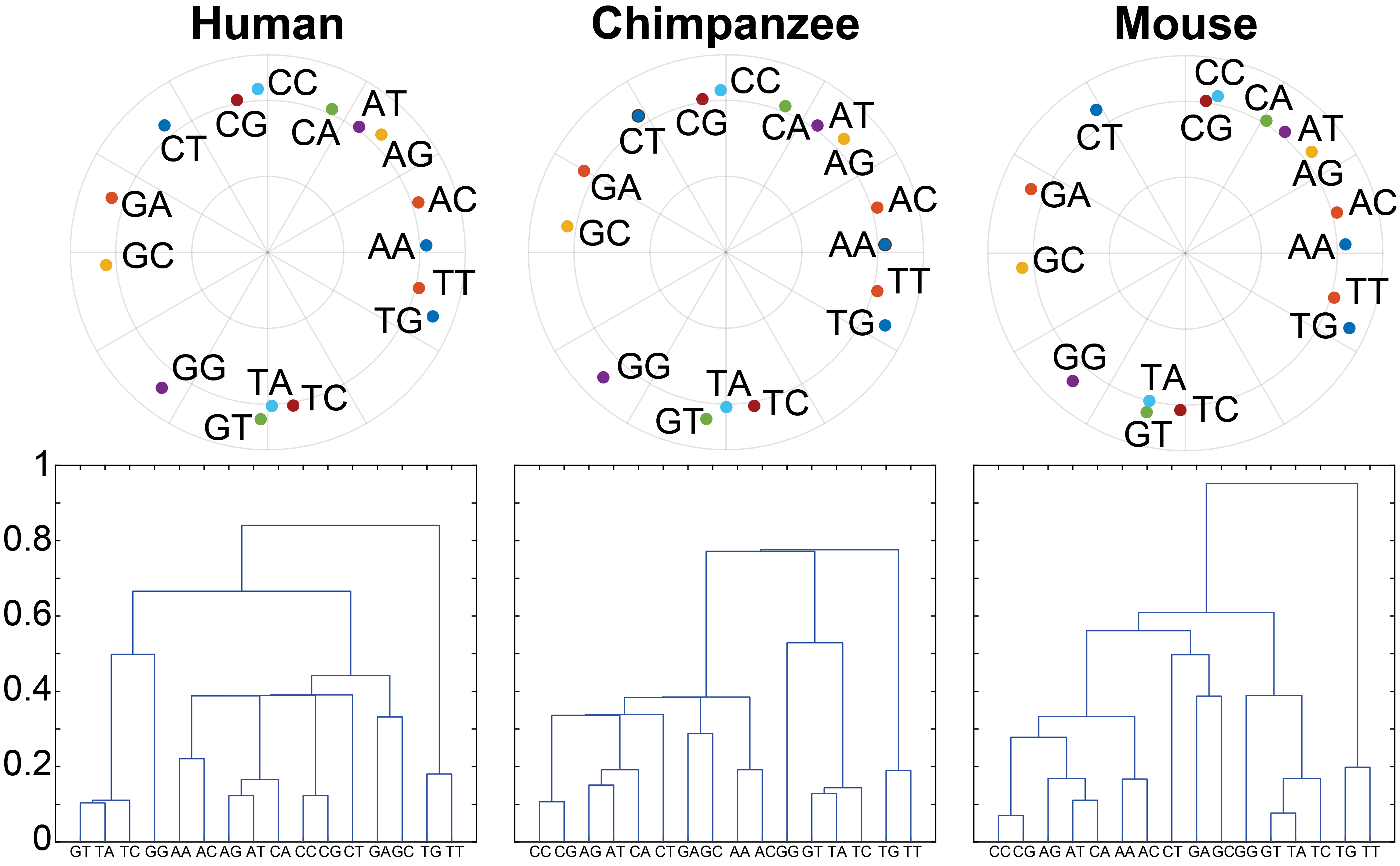}
    \caption{Hierarchical clustering result on the coding sequences of the first exon of $\beta$-globin gene of Human, Chimpanzee and Mouse. The top raw shows the polar coordinates representation of the dual nucletides. The bottom raw is the corresponding dendrogram of three species. } 
    \label{fig:dna_hier}
\end{figure}

Fig.~\ref{fig:dna_hier} presents the results of hierarchical clustering analysis on the first exon of $\beta$-globin gene of Human, Chimpanzee and Mouse~\cite{jafarzadeh2015new}. To compare the similarity between the dendrograms obtained for the three species, we computed the cophenetic correlation matrix~\cite{sokal1962comparison}. The result revealed that the dendrograms for Chimpanzee and Mouse are both similar to Human and there is a higher degree of similarity between the dendrogram for Human and Chimpanzee (cophenetic correlation: 0.944) than Human and Mouse (cophenetic correlation: 0.927, see Fig.~\ref{fig:supp_corr} in Appendix~\ref{sec:dna_seq}). These findings were consistent with the results obtained using other metrics~\cite{dai2012sequence,jafarzadeh2015new}.   

\section{Discussion}
Analysis of circular data is of increasing interest in multiple field in the life sciences. Due to the nonequivalent representation caused by converting polar coordinates to Cartesian coordinates (e.g., $\pi/2 \neq 3\pi/2$ but $\cos(\pi/2) = \cos(3\pi/2)$), most traditional clustering algorithms fail if the difference in $\theta$ plays an important role. Several circular clustering algorithms have been proposed to address this issue, but they also have disadvantages, such as being angle-focused or lacking  generality \cite{merrell2016clustering,li2018adaptive,patil2019clustering,debnath2021fast,wang2021adaptive}. Our proposed method addresses these problems by adapting the cylindrical coordinate system~\cite{roy2016swgmm,roy2017jclmm} and taking advantage of fundamental mathematical properties of polar coordinates that have been observed and successfully applied in phase-controlled coordinate choice optimization \cite{chong2018coordination,lin2020optimizing}. 

In the proposed coordinate reconstruction, varying the intermediary size of the cylindrical representation (through parameter $R$) helps to assign different weights to $\theta$ based on the research problem. With enough period repetitions, we can always find the accurate clustering result within the given dataset. Furthermore, our method is able to handle large sample sizes and multiple target classes, simply by applying a greater number of period repetitions. The computational complexity of the proposed method is relatively low, being linear in time as one increases the repetition number. The reconstruction is designed to be general for a polar coordinate system without any embedding context, so it can be broadly applied to circular clustering problems on different representations and in different fields~\cite{hussein2013assessment,li2018adaptive,rijo2019genomics,debnath2021fast}.   

\section{Conclusion}
In this paper, we propose a technique to reconstruct polar coordinates with appropriate period repetitions to form a better representation for more accurate circular clustering. This novel tool for analysis of  circular data is useful, accurate, and generally applicable. Most importantly, it can be easily interpreted and modified for various analysis needs. 


%
%


\appendices
\section{}\label{sec:heir}
\begin{figure}[h!]
    \centering
    \includegraphics[width=1\linewidth]{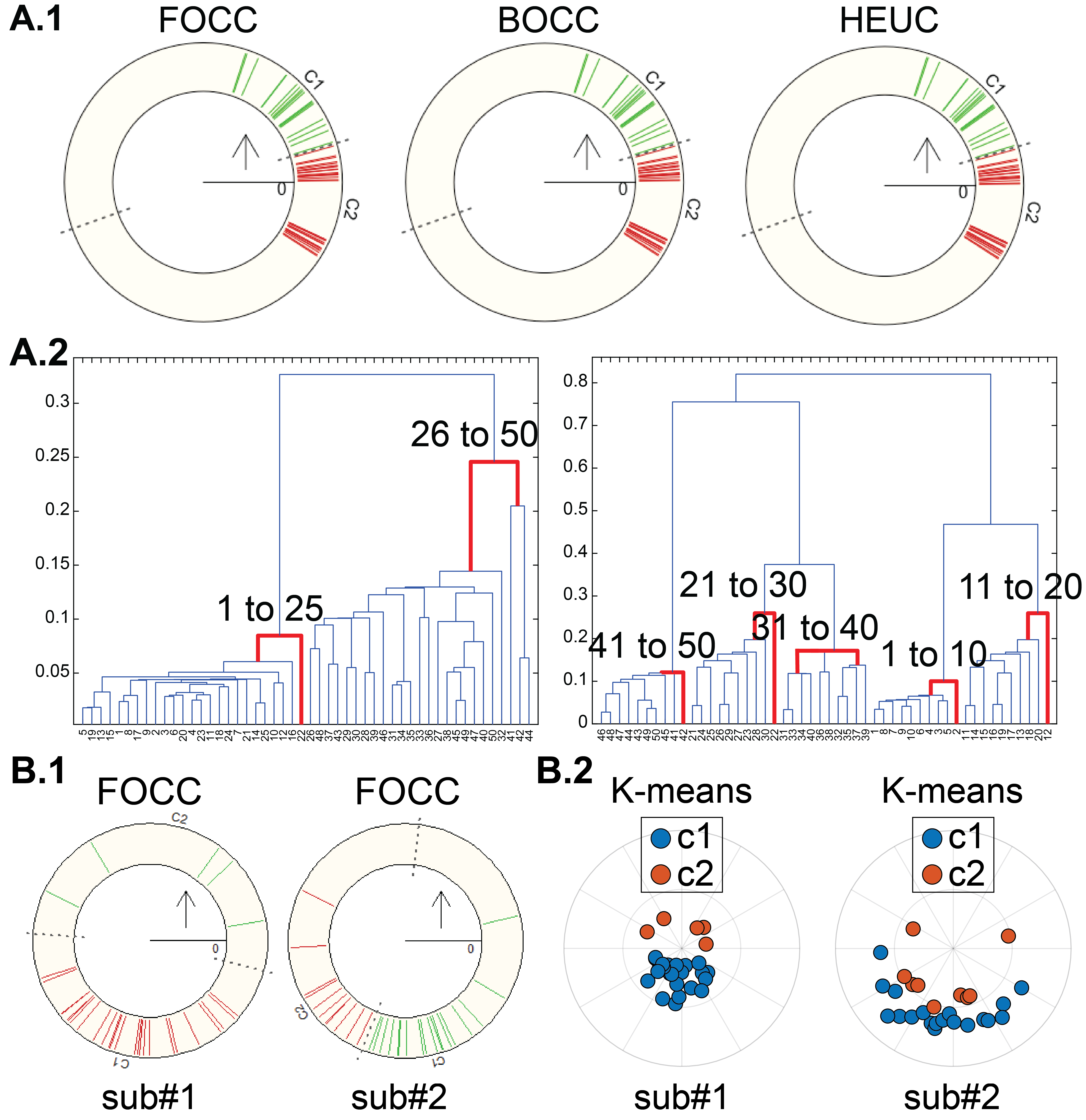}
    \caption{Additional clustering results on synthetic data. (A.1) Clustering result of FOCC and BOCC and HEUC methods on synthetic data in Section~\ref{sec:large}. All methods produce the same clustering results as our proposed method. (A.2) Hierarchical clustering results of synthetic data from Sections~\ref{sec:large} and ~\ref{sec:multi}. The same clusters are formed as $K$-means and DBSCAN (two clades (left) and five clades (right) are highlighted in red) with a balanced $R$. (B.1) FOCC method generates the same result as the angle-focused $K$-means for both subjects in Fig.~\ref{fig:realdata}. (B.2) Balanced $K$-means results for both subjects are shown, where the result of sub\#1 is the same as angle-focused $K$-means clustering result, and both subjects' results are similar to the outputs of DBSCAN method. } 
    \label{fig:supp_synth}
\end{figure}

This section presents additional clustering results of the synthetic and neural data. After applying FOCC and BOCC and HEUC -- three angle-focused circular clustering methods -- on the synthetic dataset discussed in Section~\ref{sec:large}, all methods produce identical clustering results as the proposed `angle-driven' $K$-means clustering result in Fig.~\ref{fig:synthetic}A (see Fig.~\ref{fig:supp_synth}A.1). The hierarchical clustering results for the synthetic dataset generated in Section~\ref{sec:large} and Section~\ref{sec:multi} are presented in Fig.~\ref{fig:supp_synth}A.2, where two clades are formed as the `balanced' $K$-means clustering result in Fig.~\ref{fig:synthetic}A and five clades are formed as $K$-means and DBSCAN clustering results in Fig.~\ref{fig:synthetic}B. 

Regarding the neural data shown in Fig.~\ref{fig:realdata}, the FOCC method yields the same clustering result as the angle-focused $K$-means method proposed in this study (see Fig.~\ref{fig:supp_synth}B.1). When a small $R$ (i.e., balanced weight between $r$ and $\theta$) is applied, the clustering result of sub\#1 is still the same as the angle-focused $K$-means result, since most points are clustered within a similar $\theta$ of similar $r$. This conclusion is consistent with that of the DBSCAN method. However, because sub\#2's data is more sparse, the balanced $K$-means result is different from the angle-focused result. Nonetheless, the balanced $K$-means result becomes closer to the clustering result generated by the DBSCAN method, since a balanced weight between $r$ and $\theta$ evaluates the distribution of sample points in a similar fashion to the DBSCAN method. (as the result shown in Fig.~\ref{fig:synthetic}B).              

\section{}\label{sec:dna_seq}
\begin{figure}[h!]
    \centering
    \includegraphics[width=1\linewidth]{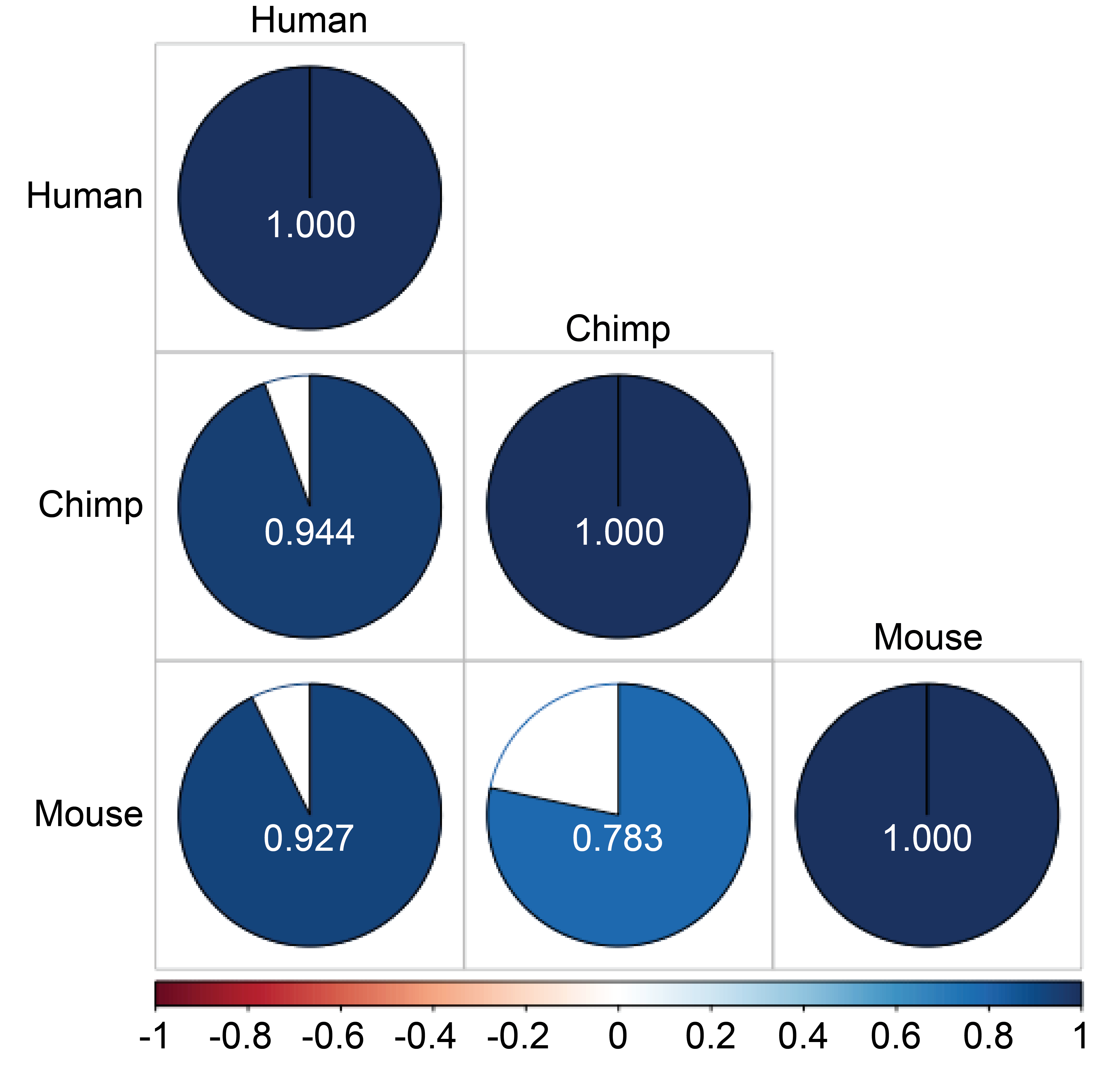}
    \caption{Cophenetic correlation matrix among hierarchical clustering result of the first exon of $\beta$-globin gene of human, chimpanzee and mouse.}
    \label{fig:supp_corr}
\end{figure}

Cophenetic correlation calculates the correlation between two cophenetic distance matrices of the two trees~\cite{sokal1962comparison}. The value can range between -1 to 1. Values near 0  mean that the two trees are not statistically similar. Cophenetic correlation matrix among hierarchical clustering results for the first exon of the $\beta$-globin gene of human, chimpanzee and mouse is shown in Fig.~\ref{fig:supp_corr}, where the result tells us that human, chimpanzee and mouse give similar results, and the similarity between human and chimpanzee is higher than human and mouse.

\ifCLASSOPTIONcompsoc
  \section*{software availability}
\else
  \section*{software availability}
\fi
All presented and evaluated algorithms are implemented in Matlab2022b (\href{https://github.com/XiaoxiaoSun0321/circular-clustering.git}{Github}).  Fig.~\ref{fig:supp_synth}A.1, Fig.~\ref{fig:supp_synth}B.1, and Fig.~\ref{fig:supp_corr} are visualized with R programming languages (package `OptCirClust':~\url{https://cran.r-project.org/web/packages/OptCirClust/index.html} and `dendrogram':~\url{https://CRAN.R-project.org/package=dendextend} ). 

\ifCLASSOPTIONcompsoc
  \section*{Acknowledgments}
\else
  \section*{Acknowledgment}
\fi
This work was funded by the National Institute of Mental Health (MH106775), a Vannevar Bush Faculty Fellowship from the US Department of Defense (N00014-20-1-2027), and a Center of Excellence grant from the Air Force Office of Scientific Research (FA9550-22-1-0337). We would like to thank Sharath Koorathota for providing feedback on the manuscript draft. We would like to thank our collaborators at Medical University of South Carolina for their help with data collection of our phase synchronization study.

\ifCLASSOPTIONcaptionsoff
  \newpage
\fi



%
\bibliographystyle{ieeetr}
\bibliography{references}

\begin{thebibliography}{10}

\bibitem{anand2015effect}
S.~Anand, P.~Padmanabham, and A.~Govardhan, ``Effect of distance measures on
  partitional clustering algorithms using transportation data,'' {\em
  International Journal of Computer Science and Information Technologies},
  vol.~6, no.~6, pp.~5308--5312, 2015.

\bibitem{merrell2016clustering}
R.~Merrell and D.~Diaz, ``Clustering analyses methods: strategies and
  algorithms,'' {\em Reviews in Theoretical Science}, vol.~4, no.~2,
  pp.~153--158, 2016.

\bibitem{patil2019clustering}
Y.~S. Patil and M.~R. Joshi, ``Clustering with polar coordinates system:
  Exploring possibilities,'' in {\em Smart Intelligent Computing and
  Applications}, pp.~553--560, Springer, 2019.

\bibitem{jain1999data}
A.~K. Jain, M.~N. Murty, and P.~J. Flynn, ``Data clustering: a review,'' {\em
  ACM computing surveys (CSUR)}, vol.~31, no.~3, pp.~264--323, 1999.

\bibitem{fell2011role}
J.~Fell and N.~Axmacher, ``The role of phase synchronization in memory
  processes,'' {\em Nature reviews neuroscience}, vol.~12, no.~2, pp.~105--118,
  2011.

\bibitem{canavier2015phase}
C.~C. Canavier, ``Phase-resetting as a tool of information transmission,'' {\em
  Current opinion in neurobiology}, vol.~31, pp.~206--213, 2015.

\bibitem{frohlich2010endogenous}
F.~Fr{\"o}hlich and D.~A. McCormick, ``Endogenous electric fields may guide
  neocortical network activity,'' {\em Neuron}, vol.~67, no.~1, pp.~129--143,
  2010.

\bibitem{qasim2021phase}
S.~E. Qasim, I.~Fried, and J.~Jacobs, ``Phase precession in the human
  hippocampus and entorhinal cortex,'' {\em Cell}, vol.~184, no.~12,
  pp.~3242--3255, 2021.

\bibitem{pantazatos2022functional}
S.~P. Pantazatos, J.~R. McIntosh, G.~T. Saber, X.~Sun, J.~Doose, J.~Faller,
  Y.~Lin, J.~B. Teves, A.~Blankenship, S.~Huffman, {\em et~al.}, ``Functional
  and effective connectivity between dorsolateral prefrontal and subgenual
  anterior cingulate cortex depends on the timing of transcranial magnetic
  stimulation relative to the phase of prefrontal alpha eeg,'' {\em bioRxiv},
  2022.

\bibitem{kristensen2019biogenesis}
L.~S. Kristensen, M.~S. Andersen, L.~V. Stagsted, K.~K. Ebbesen, T.~B. Hansen,
  and J.~Kjems, ``The biogenesis, biology and characterization of circular
  rnas,'' {\em Nature Reviews Genetics}, vol.~20, no.~11, pp.~675--691, 2019.

\bibitem{kim2020extrachromosomal}
H.~Kim, N.-P. Nguyen, K.~Turner, S.~Wu, A.~D. Gujar, J.~Luebeck, J.~Liu,
  V.~Deshpande, U.~Rajkumar, S.~Namburi, {\em et~al.}, ``Extrachromosomal dna
  is associated with oncogene amplification and poor outcome across multiple
  cancers,'' {\em Nature genetics}, vol.~52, no.~9, pp.~891--897, 2020.

\bibitem{dong2019novel}
R.~Dong, L.~He, R.~L. He, and S.~S.-T. Yau, ``A novel approach to clustering
  genome sequences using inter-nucleotide covariance,'' {\em Frontiers in
  Genetics}, vol.~10, p.~234, 2019.

\bibitem{govek2019clustering}
K.~W. Govek, V.~S. Yamajala, and P.~G. Camara, ``Clustering-independent
  analysis of genomic data using spectral simplicial theory,'' {\em PLoS
  computational biology}, vol.~15, no.~11, p.~e1007509, 2019.

\bibitem{debnath2021fast}
T.~Debnath and M.~Song, ``Fast optimal circular clustering and applications on
  round genomes,'' {\em IEEE/ACM Transactions on Computational Biology and
  Bioinformatics}, vol.~18, no.~6, pp.~2061--2071, 2021.

\bibitem{hussein2013assessment}
S.~E. Hussein, O.~A. Hassan, and M.~H. Granat, ``Assessment of the potential
  iridology for diagnosing kidney disease using wavelet analysis and neural
  networks,'' {\em Biomedical Signal Processing and Control}, vol.~8, no.~6,
  pp.~534--541, 2013.

\bibitem{mardia2000directional}
K.~V. Mardia, P.~E. Jupp, and K.~Mardia, {\em Directional statistics}, vol.~2.
\newblock Wiley Online Library, 2000.

\bibitem{charalampidis2005modified}
D.~Charalampidis, ``A modified k-means algorithm for circular invariant
  clustering,'' {\em IEEE transactions on pattern analysis and machine
  intelligence}, vol.~27, no.~12, pp.~1856--1865, 2005.

\bibitem{chang2012mean}
S.-J. Chang-Chien, W.-L. Hung, and M.-S. Yang, ``On mean shift-based clustering
  for circular data,'' {\em Soft Computing}, vol.~16, pp.~1043--1060, 2012.

\bibitem{abraham2013unsupervised}
C.~Abraham, N.~Molinari, and R.~Servien, ``Unsupervised clustering of
  multivariate circular data,'' {\em Statistics in medicine}, vol.~32, no.~8,
  pp.~1376--1382, 2013.

\bibitem{li2018adaptive}
M.~Li, M.~Stolz, Z.~Feng, M.~Kunert, R.~Henze, and F.~K{\"u}{\c{c}}{\"u}kay,
  ``An adaptive 3d grid-based clustering algorithm for automotive high
  resolution radar sensor,'' in {\em 2018 IEEE International Conference on
  Vehicular Electronics and Safety (ICVES)}, pp.~1--7, IEEE, 2018.

\bibitem{wang2021adaptive}
F.~Wang, Y.~Xie, Z.~Hu, K.~Zhang, and Y.~Zhang, ``An adaptive clustering
  algorithm based on circular units,'' in {\em 2021 4th International
  Conference on Artificial Intelligence and Big Data (ICAIBD)}, pp.~179--186,
  IEEE, 2021.

\bibitem{lin2020optimizing}
B.~Lin, B.~Chong, Y.~Ozkan-Aydin, E.~Aydin, H.~Choset, D.~I. Goldman, and
  G.~Blekherman, ``Optimizing coordinate choice for locomotion systems with
  toroidal shape spaces,'' in {\em 2020 IEEE/RSJ International Conference on
  Intelligent Robots and Systems (IROS)}, pp.~7501--7506, IEEE, 2020.

\bibitem{faller2022daily}
J.~Faller, J.~Doose, X.~Sun, J.~R. Mclntosh, G.~T. Saber, Y.~Lin, J.~B. Teves,
  A.~Blankenship, S.~Huffman, R.~I. Goldman, {\em et~al.}, ``Daily prefrontal
  closed-loop repetitive transcranial magnetic stimulation (rtms) produces
  progressive eeg quasi-alpha phase entrainment in depressed adults,'' {\em
  Brain Stimulation}, vol.~15, no.~2, pp.~458--471, 2022.

\bibitem{sun2023daily}
X.~Sun, J.~Doose, J.~Faller, J.~Mclntosh, G.~Saber, Y.~Lin, J.~Teves,
  A.~Blankenship, S.~Huffman, R.~Goldman, {\em et~al.}, ``Daily prefrontal
  closed-loop repetitive transcranial magnetic stimulation (rtms) produces a
  progressive entrainment-dependent clinical response in depressed adults,''
  {\em Bulletin of the American Physical Society}, 2023.

\bibitem{qi2009classification}
Z.-H. Qi, J.-M. Wang, and X.-Q. Qi, ``Classification analysis of dual
  nucleotides using dimension reduction,'' {\em Journal of theoretical
  biology}, vol.~260, no.~1, pp.~104--109, 2009.

\bibitem{dai2012sequence}
Q.~Dai, X.~Guo, and L.~Li, ``Sequence comparison via polar coordinates
  representation and curve tree,'' {\em Journal of theoretical biology},
  vol.~292, pp.~78--85, 2012.

\bibitem{jafarzadeh2015new}
N.~Jafarzadeh and A.~Iranmanesh, ``A new measure for pairwise comparison of
  protein sequences,'' {\em MATCH: Communications in Mathematical and in
  Computer Chemistry}, vol.~74, pp.~563--574, 2015.

\bibitem{sokal1962comparison}
R.~R. Sokal and F.~J. Rohlf, ``The comparison of dendrograms by objective
  methods,'' {\em Taxon}, pp.~33--40, 1962.

\bibitem{roy2016swgmm}
A.~Roy, S.~K. Parui, and U.~Roy, ``Swgmm: a semi-wrapped gaussian mixture model
  for clustering of circular--linear data,'' {\em Pattern Analysis and
  Applications}, vol.~19, pp.~631--645, 2016.

\bibitem{roy2017jclmm}
A.~Roy, A.~Pal, and U.~Garain, ``Jclmm: A finite mixture model for clustering
  of circular-linear data and its application to psoriatic plaque
  segmentation,'' {\em Pattern recognition}, vol.~66, pp.~160--173, 2017.

\bibitem{chong2018coordination}
B.~Chong, Y.~O. Aydin, C.~Gong, G.~Sartoretti, Y.~Wu, J.~M. Rieser, H.~Xing,
  J.~W. Rankin, K.~Michel, A.~G. Nicieza, {\em et~al.}, ``Coordination of back
  bending and leg movements for quadrupedal locomotion.,'' in {\em Robotics:
  Science and Systems}, vol.~20, Pittsburgh, PA, 2018.

\bibitem{rijo2019genomics}
F.~Rijo-Ferreira and J.~S. Takahashi, ``Genomics of circadian rhythms in health
  and disease,'' {\em Genome medicine}, vol.~11, pp.~1--16, 2019.

\end{thebibliography}

%







\end{document}